\documentclass[a4paper,fleqn]{cas-dc}

\usepackage[numbers]{natbib}
\usepackage{amsmath}
\usepackage{graphicx}
\usepackage{pifont}
\usepackage{listings}
\usepackage[most]{tcolorbox}

\newtcblisting{promptbox}[1]{
  enhanced,
  breakable,
  colback=gray!3,
  colframe=black!60,
  title=#1,
  fonttitle=\bfseries,
  listing only,
  listing options={
    basicstyle=\ttfamily\scriptsize,
    breaklines=true,
    breakatwhitespace=false,
    columns=fullflexible,
    keepspaces=true,
    showstringspaces=false
  },
  left=1mm,
  right=1mm,
  top=1mm,
  bottom=1mm,
  boxsep=1mm,
  arc=1mm,
  before skip=-2pt,
  after skip=4pt
}

\def\tsc#1{\csdef{#1}{\textsc{\lowercase{#1}}\xspace}}
\tsc{WGM}
\tsc{QE}

\begin{document}
\let\WriteBookmarks\relax
\def\floatpagepagefraction{1}
\def\textpagefraction{.001}

\shorttitle{}    

\shortauthors{}  

\title [mode = title]{Training-Free Dense Hand Contact Estimation with Multi-Modal Large Language Models}  

\author[1]{Daniel Sungho Jung}[orcid=0000-0002-9955-418X]
\ead{dqj5182@snu.ac.kr}
\credit{Conceptualization, Methodology, Software, Validation, Investigation, Visualization, Writing}

\affiliation[1]{organization={Interdisciplinary Program in Artificial Intelligence, Seoul National University},
            city={Seoul},
            postcode={08826},
            country={Republic of Korea}}

\author[1,2]{Kyoung Mu Lee}[orcid=0000-0001-7210-1036]
\cormark[1]
\ead{kyoungmu@snu.ac.kr}
\credit{Supervision, Writing}

\affiliation[2]{organization={Department of Electrical and Computer Engineering and ASRI, Seoul National University},
            city={Seoul},
            postcode={08826},
            country={Republic of Korea}}

\cortext[1]{Corresponding author}

\begin{abstract}
Dense hand contact estimation requires both high-level semantic understanding and fine-grained geometric reasoning of human interaction to accurately localize contact regions. 
Recently, multi-modal large language models (MLLMs) have demonstrated strong capabilities in understanding visual semantics, enabled by vision–language priors learned from large-scale data. 
However, leveraging MLLMs for dense hand contact estimation remains underexplored. 
There are two major challenges in applying MLLMs to dense hand contact estimation. 
First, encoding explicit 3D hand geometry is difficult, as MLLMs primarily operate on vision and language modalities. 
Second, capturing fine-grained vertex-level contact remains challenging, as MLLMs tend to focus on high-level semantics rather than detailed geometric reasoning. 
To address these challenges, we propose ContactPrompt, a training-free and zero-shot approach for dense hand contact estimation using MLLMs. 
To effectively encode 3D hand geometry, we introduce a detailed hand-part segmentation and a part-wise vertex-grid representation that provide structured, localized geometric information. 
To enable accurate and efficient dense contact prediction, we develop a multi-stage contact reasoning with part conditioning, progressively bridging global semantics and fine-grained geometry. 
Therefore, our method effectively leverages the reasoning capabilities of MLLMs to enable precise dense hand contact estimation. 
Surprisingly, the proposed approach achieves performance comparable to the previous state-of-the-art method trained on large-scale dense contact datasets, without requiring any task-specific training.
\end{abstract}


\maketitle

\section{Introduction}
\label{sec:intro}

From everyday object manipulation to complex tasks, humans interact with the world through their hands, guided by semantic intentions shaped by language-based reasoning.
Most hand actions are driven by such intentions, reflecting underlying semantic meaning, such as holding a cup or pressing a button, which can be naturally expressed in language.
Accordingly, developing a dense hand contact estimation model that effectively leverages the semantic meaning of human interaction is essential for accurate, semantically plausible hand contact prediction.

Recently, multi-modal large language models (MLLMs) \cite{openai2026gpt54, openai2026gpt55, anthropic2026claude_sonnet46, anthropic2026claude_opus_47} exhibit remarkable performance across a wide range of tasks, driven by powerful language-based reasoning combined with predominantly visual multi-modal inputs.
Prior works have successfully leveraged MLLMs as high-level semantic guidance for vision tasks~\cite{yu2024scaling} or as auxiliary modules to improve generalization~\cite{badalyan2026ngl}.
Nevertheless, despite these promising results, leveraging MLLMs for 3D reasoning tasks remains underexplored due to the difficulty of directly encoding explicit 3D geometric representations (\textit{e.g.,} meshes, point clouds) and the challenge of predicting fine-grained 3D geometry.
In this paper, we aim to develop a framework that directly leverages the power of MLLMs for both high-level semantic understanding and fine-grained geometric reasoning in dense hand contact estimation.

There are two major challenges that must be addressed to effectively leverage MLLM capabilities for dense hand contact estimation.
First, directly encoding the 3D geometry of the human hand is difficult, as MLLMs primarily operate on vision and language modalities.
A straightforward approach to providing 3D geometry is to supply raw 3D mesh data of the MANO hand model~\cite{romero2017embodied} to MLLMs.
However, most MLLMs convert such 3D mesh data into text and process the geometry as textual input.
As MLLMs are not designed to analyze 3D coordinates and their spatial relationships, they often fail to capture the underlying 3D structure of the human hand when provided with raw geometric data.
Second, capturing fine-grained vertex-level contact from images with MLLMs remains limited, as they primarily focus on high-level semantic reasoning unless provided with specific prompts or guidance that precisely define and describe each hand vertex.
Providing a text prompt for each vertex of the MANO hand model requires 778 sentences corresponding to its 778 vertices, resulting in excessively long inputs that are inefficient to process with MLLMs.
Even when such prompts are made concise, constructing descriptions that can distinguish between closely positioned vertices within the hand mesh remains challenging, as language is inherently ambiguous for fine-grained spatial reasoning.
Therefore, developing an effective representation and reasoning framework at the vertex-level remains underexplored yet essential for fully leveraging MLLMs for dense hand contact estimation.

To tackle these issues, we propose ContactPrompt, a framework for dense hand contact estimation that enables MLLMs to effectively utilize their high-level semantic understanding capability for fine-grained geometric reasoning. 
Instead of directly providing raw 3D geometry to MLLMs, ContactPrompt introduces a structured geometric representation that makes 3D hand geometry interpretable to MLLMs via visual prompts. 
Specifically, we first define a detailed hand-part segmentation that decomposes the hand into fine-grained, functionally meaningful regions. 
Based on this segmentation, we construct a part-wise vertex-grid representation that organizes hand vertices into structured grids, enabling localized, spatially coherent reasoning by MLLMs.
To facilitate accurate and efficient dense contact prediction, we build upon the structured geometric representation and formulate the prediction task as a multi-stage reasoning process, where the model progressively refines predictions from global interaction to part-level contact and finally to dense vertex-level contact. 
To further improve efficiency and prediction focus, we introduce part conditioning, which restricts dense prediction to the most relevant hand part regions. 
Through this formulation, ContactPrompt enables MLLMs to bridge global semantic understanding and fine-grained geometric prediction, achieving accurate dense hand contact estimation without any task-specific training.

As a result, ContactPrompt achieves performance comparable to the previous state-of-the-art method trained on large-scale dense contact datasets without requiring any task-specific training. 
Our key contributions are as follows:
\begin{itemize}
    \item We introduce ContactPrompt, a novel, training-free, zero-shot framework that enables MLLMs to perform dense hand contact estimation.
    \item To encode 3D hand geometry for MLLMs, we present a detailed hand-part segmentation and a part-wise vertex-grid representation that enables structured encoding of 3D hand geometry for MLLM reasoning.
    \item To enable accurate and efficient dense hand contact estimation, we develop a multi-stage contact reasoning with part conditioning, which progressively bridges global semantic understanding of MLLMs and fine-grained geometric prediction.
    \item In the end, ContactPrompt achieves performance comparable to the state-of-the-art method trained on large-scale dense contact datasets without requiring any task-specific training.
\end{itemize}
\section{Related works}
\label{sec:related_works}

\noindent\textbf{Dense hand contact estimation.}
Most existing methods for dense hand contact estimation rely on task-specific datasets~\cite{tzionas2016capturing, cao2021reconstructing, huang2022capturing, yin2023hi4d} that provide dense contact labels. 
POSA~\cite{hassan2021populating} models contact probability conditioned on 3D body pose using a cVAE framework. 
BSTRO~\cite{huang2022capturing} leverages a Transformer-based architecture to estimate dense body–scene contact by capturing non-local relationships. 
DECO~\cite{tripathi2023deco} employs cross-attention to integrate scene context and part information learned from 2D supervisions. 
HACO~\cite{jung2025learning} addresses class and spatial imbalance problems in hand contact datasets through balanced contact sampling and vertex-level class balancing loss design. 
GECO~\cite{lee2024geco} explores MLLMs for contact estimation by predicting contact in semantically defined body parts. 
Despite these advances, achieving generalizable dense hand contact estimation remains underexplored.
HACO remains limited by task-specific supervision, which causes generalization constraints. 
GECO is limited to part-level predictions and focuses on full-body contact.
In contrast, ContactPrompt enables generalizable dense hand contact estimation by formulating the task as a multi-stage reasoning problem with MLLMs, allowing fine-grained vertex-level predictions without task-specific training.

\noindent\textbf{Prompting for 3D reasoning with MLLMs.}
Recent works have explored the use of MLLMs for 3D reasoning through structured representations and prompting strategies.
SG-Nav~\cite{yin2024sg} utilizes object-level coordinates and hierarchical scene graphs for spatial reasoning, while CE3D~\cite{fang2024chat} encodes scene geometry into intermediate representation of atlas. 
Other approaches provide explicit geometric priors or cues, including 3DAxisPrompt~\cite{liu20253daxisprompt}, which uses coordinate axes and segmentation masks, and See\&Trek~\cite{lisee}, which incorporates keyframes and motion cues for trajectory reasoning. 
NGL-Prompter~\cite{badalyan2026ngl} and PromptVFX~\cite{kiraypromptvfx} further extend this direction by demonstrating the effectiveness of language-friendly representations for structured generation tasks.
Despite these advances, existing methods primarily focus on high-level spatial reasoning and do not address fine-grained geometric prediction. 
In contrast, ContactPrompt enables training-free hand contact estimation by introducing multi-stage contact reasoning, enabling MLLMs to make localized, spatially coherent vertex-level predictions. 
\section{Method}

\begin{figure*}[pos=t]
\begin{center}
\includegraphics[width=0.9\linewidth]{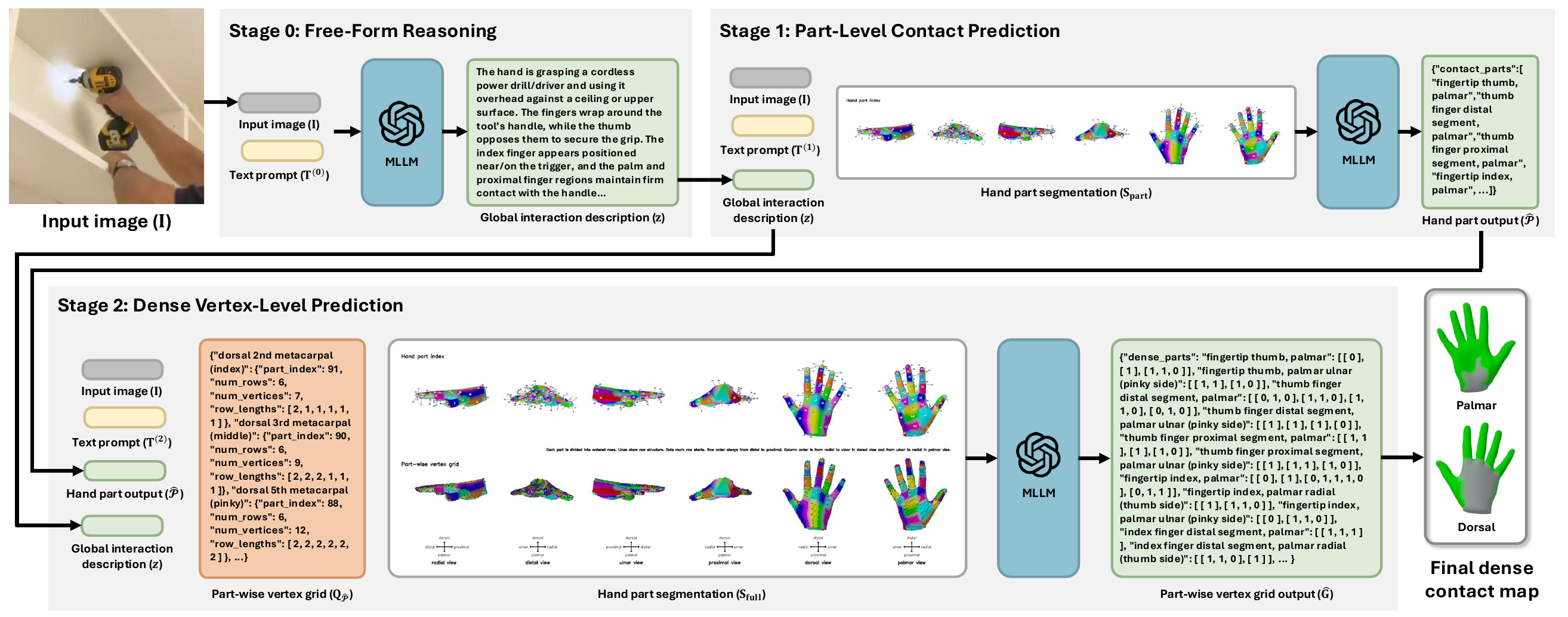}
\end{center}
\vspace{-3.5mm}
\caption{
\textbf{Overall pipeline of ContactPrompt.} Given an input image $\mathbf{I}$ and text prompt~$\mathbf{T}^{(0)}$, we first perform free-form reasoning with MLLMs to produce a global interaction description $\mathbf{z}$.
Next, part-level contact prediction is performed using $\mathbf{I}$, $\mathbf{z}$, a text prompt~$\mathbf{T}^{(1)}$, and hand-part segmentation $\mathbf{S}_{\text{part}}$ to obtain predicted contact parts $\hat{\mathcal{P}}$.
Moreover, dense vertex-level contact is estimated by providing $\mathbf{I}$, $\mathbf{T}^{(2)}$, $\hat{\mathcal{P}}$,  $\mathbf{z}$, the full visual prompt $\mathbf{S}_{\text{full}}$, and the part-wise vertex-grid specification~$\mathbf{Q}_{\hat{\mathcal{P}}}$, producing part-wise grid outputs $\hat{\mathbf{G}}$.
Lastly, the output~$\hat{\mathbf{G}}$ is mapped to the final dense hand contact map.}
\label{fig:contactprompt_overall_pipeline}
\vspace{-0.5cm}
\end{figure*}

We address dense hand contact estimation by formulating the task as a multi-stage reasoning problem with multi-modal large language models (MLLMs), as illustrated in Figure~\ref{fig:contactprompt_overall_pipeline}.
Given an input RGB image $\mathbf{I}$, our objective is to predict binary contact labels over the MANO hand mesh~\cite{romero2017embodied} with $V = 778$ vertices. 
Rather than directly regressing contact from images, we decompose the task into multiple stages that progressively bridge the global semantic understanding of MLLMs and fine-grained geometric prediction.

\subsection{Detailed hand-part segmentation}

\begin{figure}[pos=t]
\begin{center}
\includegraphics[width=1.0\linewidth]{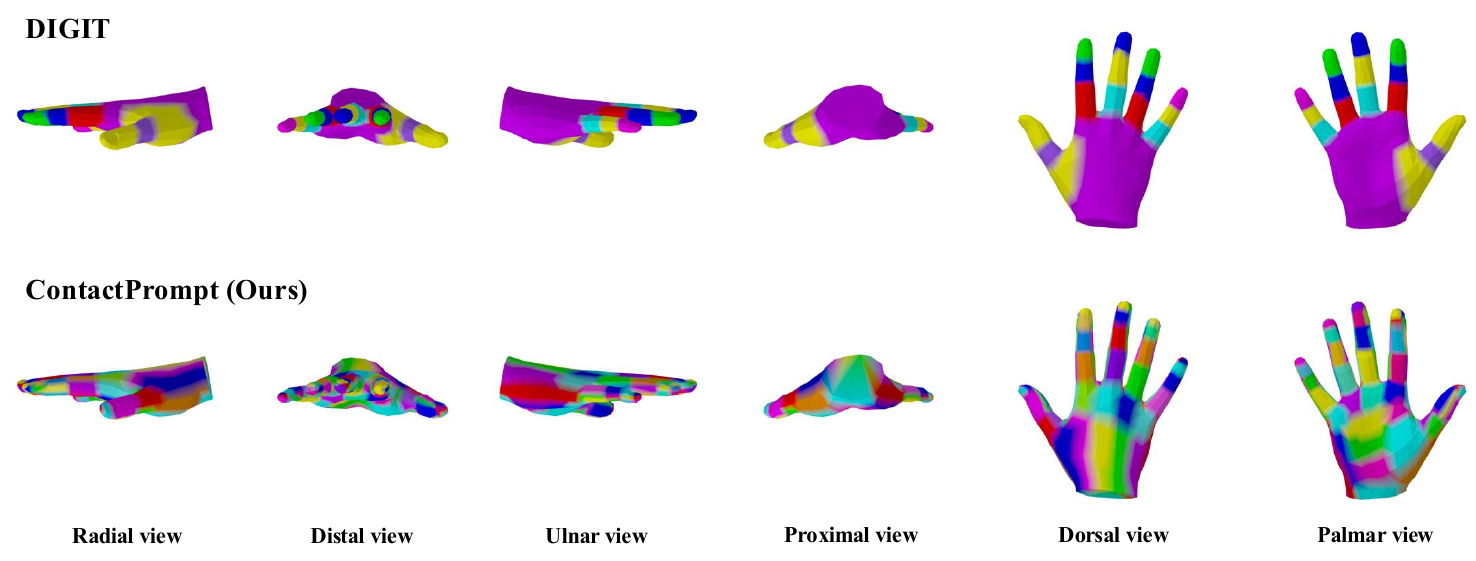}
\end{center}
\vspace{-2.5mm}
\caption{
\textbf{Comparison of hand-part segmentation with DIGIT~\cite{fan2021learning}.} Our ContactPrompt provides more detailed hand-part segmentation aligned with the function of hand parts.}
\label{fig:contactprompt_digit_comparison}
\vspace{-0.5mm}
\end{figure}

Let the MANO hand mesh be defined by vertices $\mathbf{V} \in \mathbb{R}^{V \times 3}$. 
The hand is partitioned into a set of semantic parts $\mathcal{P} = \{ p_1, p_2, \dots, p_K \}$, where each part $p$ corresponds to a subset of vertices $\mathcal{V}_p \subset \{1, \dots, V\}$. 
As shown in Figure~\ref{fig:contactprompt_digit_comparison}, our hand-part segmentation differs from prior work, such as DIGIT~\cite{fan2021learning}, by providing a more detailed, functionally aligned decomposition of the hand. 
To achieve this, the hand is first divided based on major surface orientations, including palmar, dorsal, palmar radial, and palmar ulnar. 
Palmar radial and palmar ulnar refer to the lateral regions of the hand oriented toward the thumb and pinky finger, respectively. 
Within the palmar hand body, regions are further decomposed into finger bases, multiple palm center regions spanning distal, middle, and proximal areas, thenar regions, wrist regions, and lateral hand-side regions. 
The dorsal hand body is divided into knuckle regions and metacarpal regions corresponding to each finger. 
Finger bases are defined as the palmar regions immediately below each finger and adjacent to the knuckles. 
Finger regions are segmented into proximal, intermediate, and distal segments, with orientation-specific subdivisions, as well as fingertips, which serve as representative contact regions. 
Webspace regions between adjacent fingers, especially between the thumb and index finger, are explicitly defined due to their importance in fine manipulation tasks, such as holding a pen using the thumb–index webspace. 
This detailed segmentation is designed to be visually distinguishable and semantically meaningful, enabling MLLMs to more effectively associate language-based reasoning with localized geometric regions of the hand.
In total, this segmentation defines $K = 103$ semantic hand parts. 
This level of granularity provides a strong densification of the hand representation relative to the full set of 778 MANO vertices, enabling fine-grained yet semantically grounded reasoning for dense hand contact estimation with MLLMs.

\subsection{Part-wise vertex-grid representation}
\label{sec:part-wise_vertex_grid}

To enable dense hand contact estimation, a part-wise vertex-grid representation is defined for each segmented hand part $p \in \mathcal{P}$.
The vertices of each part are organized into an ordered set of rows:
\begin{equation}
\mathcal{G}_p = \{ \mathbf{g}_p^{(1)}, \mathbf{g}_p^{(2)}, \dots, \mathbf{g}_p^{(r_p)}, \dots, \mathbf{g}_p^{(R_p-1)}, \mathbf{g}_p^{(R_p)} \},
\end{equation}
where $R_p$ denotes the number of rows for part~$p$, and $r_p \in \{1, \dots, R_p\}$ indexes each row. 
Each row $\mathbf{g}_p^{(r_p)}$ consists of an ordered list of vertices with length defined by $\texttt{row\_lengths}[r_p]$, following the predefined part-wise grid specification provided to the MLLM. 
The rows are ordered from fingertip to wrist, and the vertices within each row are arranged from left to right in the corresponding view, as illustrated in the part-wise vertex-grid of the visual prompt in Figure~\ref{fig:contactprompt_visual_prompt}. 
The visual prompt further depicts the start of each row with a dot, lines across each row, and connections between the end of one row and the start of the next, explicitly conveying the grid's sequential structure. 
This construction ensures that vertices within each row are spatially adjacent on the mesh, while consecutive rows follow the surface topology of the hand part, forming a compact, structured 2D-like layout that preserves local geometric continuity.
Based on this representation, dense hand contact is predicted in a structured form where each part name is associated with its corresponding part-wise vertex-grid, and each grid element is predicted as a binary contact value of 0 or 1. 
This prediction is enforced via text prompts that require strict adherence to the predefined grid structure. 
The part-wise vertex-grid specification is provided to the MLLM in JSON format, including the part name, part index, number of rows, row lengths, and the total number of vertices for each part. 
Explicit vertex indices for each grid element are not provided to the MLLM, as the prediction only requires binary contact assignment for each element within the part-wise vertex-grid. 
Finally, the predicted grid outputs are aggregated using the predefined part-wise vertex-grid to MANO vertex mapping to obtain the vertex-level contact vector $\mathbf{c} \in \{0,1\}^{V}$, where $\mathbf{c}$ denotes the binary contact state for all MANO vertices.

\subsection{Multi-stage contact reasoning with MLLMs}
\label{sec:multi-stage}

\begin{figure}[pos=t]
\begin{center}
\includegraphics[width=1.0\linewidth]{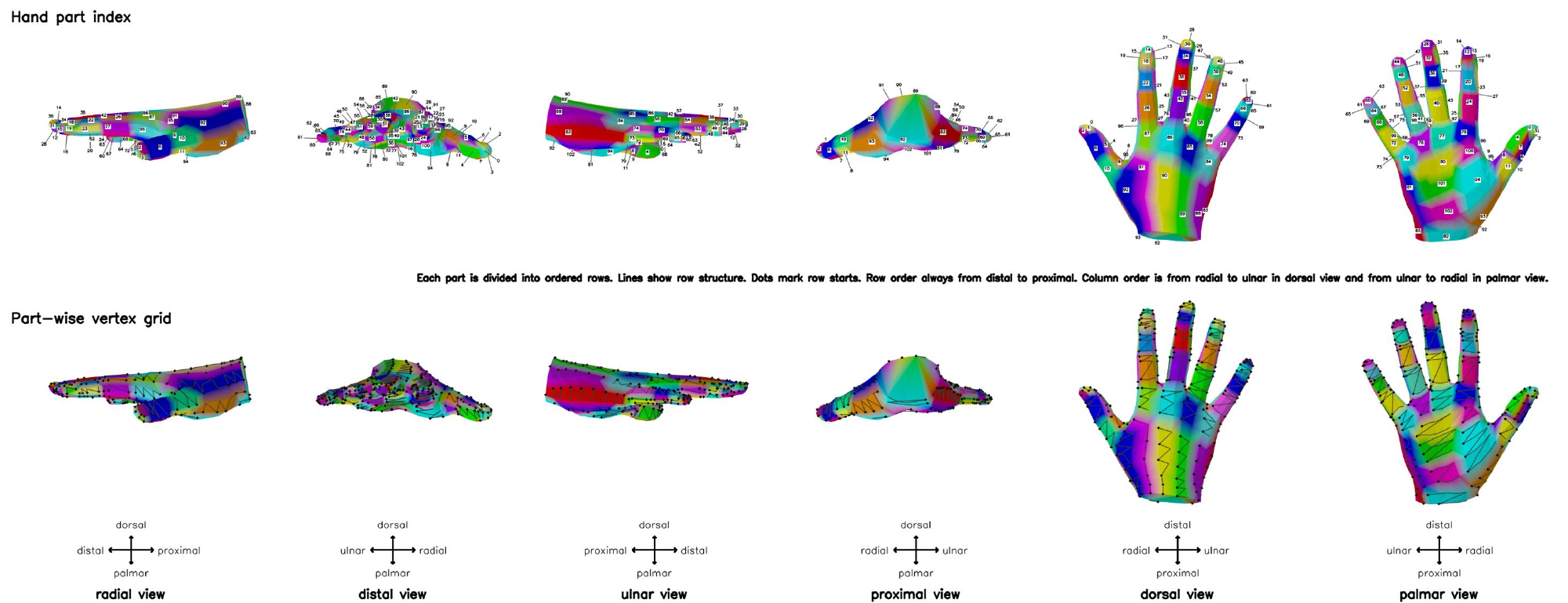}
\end{center}
\vspace{-2mm}
\caption{
\textbf{Details of visual prompt in ContactPrompt.} The visual prompt consists of hand part index and part-wise vertex-grid. 
Hand part index associates each region with numeric label. 
Vertex-grid shows row structure, where each row starts with a dot, vertices connected by lines, and consecutive rows linked to indicate sequential ordering between rows of the grid.}
\label{fig:contactprompt_visual_prompt}
\vspace{-0.5cm}
\end{figure}

Dense hand contact estimation is formulated as a multi-stage reasoning process as shown in Figure~\ref{fig:contactprompt_overall_pipeline}. 
The model operates in three stages: free-form stage~($f^{(0)}$), part-level stage~($f^{(1)}$), and dense vertex-level stage~($f^{(2)}$), each guided by stage-specific text and visual prompts. 
In the free-form stage, the MLLM generates a global interaction description:
\begin{equation}
\mathbf{z} = f^{(0)}(\mathbf{I}, \mathbf{T}^{(0)}),
\end{equation}
where $\mathbf{I}$ denotes the input RGB image and $\mathbf{T}^{(0)}$ denotes the text prompt for the free-form reasoning stage~(stage 0). 
The prompt $\mathbf{T}^{(0)}$ guides the model to reason about hand pose, camera viewpoint, object interaction, occlusion, and physically plausible contact regions. 
The output $\mathbf{z}$ is a free-form textual description capturing high-level semantic understanding of the interaction. 
In the part-level stage, the MLLM predicts which hand parts are in contact:
\begin{equation}
\hat{\mathcal{P}} = f^{(1)}(\mathbf{I}, \mathbf{T}^{(1)}, \mathbf{S}_{\text{part}}, \mathbf{z}),
\end{equation}
where $\mathbf{S}_{\text{part}}$ denotes the hand part index visual prompt shown as the upper row of Figure~\ref{fig:contactprompt_visual_prompt}, and $\mathbf{T}^{(1)}$ denotes the text prompt for the part-level contact prediction stage~(stage 1). 
The output $\hat{\mathcal{P}} = \{ p_1, \dots, p_k \}$ is a set of predicted contact hand parts, where $k$ denotes the number of predicted parts. 
This stage utilizes global reasoning $\mathbf{z}$ from the free-form stage and geometric cues derived from the visual prompt $\mathbf{S}_{\text{part}}$ to identify semantically and spatially plausible contact regions. 
In the dense vertex-level stage, dense contact is predicted only for the selected parts using part conditioning:
\begin{equation}
\hat{\mathbf{G}} = f^{(2)}(\mathbf{I}, \mathbf{T}^{(2)}, \mathbf{S}_{\text{full}}, \mathbf{z}, \hat{\mathcal{P}}, \mathbf{Q}_{\hat{\mathcal{P}}}),
\end{equation}
where $\mathbf{S}_{\text{full}}$ denotes the full visual prompt shown as Figure~\ref{fig:contactprompt_visual_prompt}, $\mathbf{T}^{(2)}$ denotes the text prompt for dense vertex-level prediction stage~(stage 2), and $\mathbf{Q}_{\hat{\mathcal{P}}}$ denotes the grid specification~$\mathbf{Q}$ for the selected parts~$\hat{\mathcal{P}}$, including the number of rows and row lengths. 
The output $\hat{\mathbf{G}} = \{ \hat{\mathbf{G}}_p \mid p \in \hat{\mathcal{P}} \}$ consists of part-wise vertex-grids, where each $\hat{\mathbf{G}}_p$ follows the predefined row structure specified by $\mathbf{Q}_{\hat{\mathcal{P}}}$.
Part conditioning restricts the prediction space to $\hat{\mathcal{P}}$, enabling more focused and efficient dense hand contact estimation. 
The final vertex-level contact prediction is obtained by aggregating the part-wise grid outputs~$\hat{\mathbf{G}}$ as described in Section~\ref{sec:part-wise_vertex_grid}. 
To ensure valid outputs, structural constraints on the part-wise vertex-grid are strictly enforced through text prompts, requiring each predicted grid to exactly match the specified number of rows and row lengths with binary values. 
Each stage allows a limited number of regenerations when outputs are invalid or incomplete.
In such cases, error feedback describing violations of structural constraints is appended to the text prompt, guiding the MLLM to self-correct its previous output.

\subsection{Efficient dense contact estimation via part conditioning}

To reduce computational overhead and improve prediction focus, we restrict dense contact estimation to the predicted contact parts. 
Let $\hat{\mathcal{P}}$ denote the set of predicted contact parts from the part-level stage, and let $\mathcal{V}_p$ denote the predefined set of vertices associated with part $p$. 
Part conditioning defines the effective prediction domain as follows:
\begin{equation}
\mathcal{V}_{\text{active}} = \bigcup_{p \in \hat{\mathcal{P}}} \mathcal{V}_p,
\end{equation}
which corresponds to the union of vertices belonging to the predicted contact parts. 
For vertices outside this set, the contact state is assigned as non-contact:
\begin{equation}
\hat{c}_v = 0, \quad \forall v \notin \mathcal{V}_{\text{active}},
\end{equation}
where $\hat{c}_v$ denotes the predicted binary contact value of vertex $v$. 
This reduces the effective prediction size from the full set of $V$ vertices to a smaller subset $V' = |\mathcal{V}_{\text{active}}|$, leading to fewer output tokens and improved inference efficiency during the dense vertex-level stage of the multi-stage contact reasoning described in Section~\ref{sec:multi-stage}.

\section{Implementation details}

GPT-5.5~\cite{openai2026gpt55} is used as the base MLLM for ContactPrompt via the OpenAI API, and all inference is performed in a training-free, zero-shot manner.
All images, including the input RGB image and visual prompts, are encoded as base64 JPEGs before being passed to the MLLM, while textual prompts are provided directly without additional preprocessing.
The contact reasoning pipeline in Section~\ref{sec:multi-stage} allows a fixed number of retries for each stage, where the part stage allows up to 2 retries and the dense vertex-level stage allows up to 4 retries when outputs are invalid, incomplete, or violate structural constraints.
We adopt the MANO hand model~\cite{romero2017embodied} with 778 vertices.
All inference is performed per sample due to the sequential dependency across stages.
Predicted contact is evaluated using a threshold of 0.5 to compute precision, recall, and F1-score.
For RICH and Hi4D, we evaluate on 100 samples uniformly subsampled from their respective test sets.
All experiments are conducted on a single NVIDIA RTX A6000 GPU for data processing, rendering, and running open-source MLLMs.
Closed-source MLLMs are accessed through external APIs, while open-source MLLMs are evaluated locally on the same GPU.
\section{Experiments}

\subsection{Datasets}
We follow HACO~\cite{jung2025learning} and use the MOW~\cite{cao2021reconstructing} dataset as the primary benchmark, as it offers diverse in-the-wild hand-object interaction scenarios with 3D annotations that best reflect real-world conditions. 
The dataset consists of 92 samples from the standard evaluation split. 
For evaluation, we use the dense hand contact annotations provided by HACO, derived from the ground-truth 3D hand and object mesh annotations.
We additionally evaluate on HIC~\cite{tzionas2016capturing}, RICH~\cite{huang2022capturing}, and Hi4D~\cite{yin2023hi4d} datasets to assess the applicability of ContactPrompt across diverse hand interaction datasets.

\subsection{Evaluation metrics}
To evaluate dense hand contact estimation, we compute precision, recall, and F1-score at the vertex-level on the MANO hand mesh~\cite{romero2017embodied}. 
In addition to contact accuracy, we evaluate MLLM inference efficiency by measuring the number of output tokens and the inference cost per sample. 
The inference cost is reported in US dollars~(\$) based on the API pricing at the time of experiments, where the OpenAI API cost is \$30.00 per 1M output tokens for GPT-5.5~\cite{openai2026gpt55} and \$15.00 per 1M output tokens for GPT-5.4~\cite{openai2026gpt54}.

\begin{table}[pos=t]
\centering
\footnotesize
\setlength{\tabcolsep}{1.5pt}
\caption{\textbf{Ablation of hand-part segmentation on MOW~\cite{cao2021reconstructing} dataset.}}
\begin{tabular}{lcccc}
\toprule
Methods & \# of parts
        & Precision~$\uparrow$ & Recall~$\uparrow$ & F1-Score~$\uparrow$ \\
\midrule
DIGIT~\cite{fan2021learning} & 16 & 0.404 & 0.464 & 0.389 \\
\midrule
ContactPrompt & 32 & 0.407 & \textbf{0.780} & 0.499 \\
ContactPrompt & 65 & \underline{0.460} & 0.707 & \underline{0.513} \\
ContactPrompt~(Ours) & 103
    & \textbf{0.473} & \underline{0.710} & \textbf{0.526} \\
\bottomrule
\end{tabular}
\label{tab:abl_hand_part_seg}
\end{table}

\begin{table}[pos=t]
\centering
\footnotesize
\setlength{\tabcolsep}{2pt}
\caption{\textbf{Ablation of part-wise vertex-grid on MOW~\cite{cao2021reconstructing} dataset.}}
\begin{tabular}{ccccc}
\toprule
Part-wise & Vertex-grid
& Precision~$\uparrow$ & Recall~$\uparrow$ & F1-Score~$\uparrow$ \\
\midrule
\ding{55} & \ding{55} & 0.288 & 0.278 & 0.261 \\
\ding{51} & \ding{55} & \textbf{0.508} & \underline{0.456} & \underline{0.432} \\
\ding{51} & \ding{51} & \underline{0.473} & \textbf{0.710} & \textbf{0.526} \\
\bottomrule
\end{tabular}
\label{tab:abl_part_grid}
\end{table}

\begin{table}[pos=t]
\centering
\footnotesize
\setlength{\tabcolsep}{2pt}
\caption{\textbf{Ablation of geometric representation on MOW~\cite{cao2021reconstructing} dataset.}}
\begin{tabular}{lccc}
\toprule
Representation & Precision~$\uparrow$ & Recall~$\uparrow$ & F1-Score~$\uparrow$ \\ 
\midrule 
3D vertex coordinates & \textbf{0.529} & 0.498 & 0.454 \\ 
Part-wise vertex-grid~(Ours) & 0.473 & \textbf{0.710} & \textbf{0.526} \\
\bottomrule
\end{tabular}
\label{tab:abl_part_wise_vertex_grid_geo}
\end{table}

\subsection{Ablation studies}

\noindent\textbf{Effectiveness of detailed hand-part segmentation.}
In Table~\ref{tab:abl_hand_part_seg}, the detailed hand-part segmentation improves contact accuracy.
Compared to DIGIT~\cite{fan2021learning}, our method improves F1-score by 35.2\%. 
Increasing the granularity of the segmentation improves F1-score from 0.499 with 32 parts to 0.513 with 65 parts and 0.526 with 103 parts, demonstrating the benefit of fine-grained semantic decomposition.

\noindent\textbf{Effectiveness of part-wise vertex-grid.}
In Table~\ref{tab:abl_part_grid}, we separately analyze the effects of part-wise decomposition and vertex-grid organization. 
Organizing the whole hand as a single 1D vertex sequence~(first row of Table~\ref{tab:abl_part_grid}) results in an F1-score of 0.261.
Decomposing into hand parts but with single 1D vertex sequence for each part~(second row of Table~\ref{tab:abl_part_grid}) substantially improves the F1-score to 0.432, demonstrating the importance of part-wise reasoning. 
Further organizing the vertices within each part into the proposed vertex-grid increases the F1-score to 0.526. 
Overall, these results demonstrate that the part-wise vertex-grid plays a critical role in enhancing dense contact estimation.

\begin{table}[pos=t]
\centering
\footnotesize
\setlength{\tabcolsep}{2pt}
\caption{\textbf{Ablation of multi-stage contact reasoning on MOW~\cite{cao2021reconstructing} dataset.}}
\begin{tabular}{cccccc}
\toprule
Free-form & Part & Dense 
& Precision~$\uparrow$ & Recall~$\uparrow$ & F1-Score~$\uparrow$ \\ 
\midrule 
\ding{55} & \ding{55} & \ding{51} & 0.382 & \textbf{0.944} & \underline{0.513} \\ 
\ding{55} & \ding{51} & \ding{51} & \underline{0.460} & 0.658 & 0.497 \\ 
\ding{51} & \ding{55} & \ding{51} & 0.379 & 0.669 & 0.435 \\ 
\ding{51} & \ding{51} & \ding{51} & \textbf{0.473} & \underline{0.710} & \textbf{0.526} \\ 
\bottomrule
\end{tabular}
\label{tab:abl_multi_stage}
\end{table}

\noindent\textbf{Effectiveness of geometric representation for dense contact.}
In Table~\ref{tab:abl_part_wise_vertex_grid_geo}, we compare the proposed part-wise vertex-grid with directly providing 3D vertex coordinates and their corresponding vertex indices for dense vertex-level prediction (stage 2).
The proposed part-wise vertex-grid improves F1-score from 0.454 to 0.526, demonstrating the benefit
of an MLLM-friendly geometric representation.

\begin{table}[pos=t]
\centering
\scriptsize
\setlength{\tabcolsep}{2pt}
\caption{\textbf{Ablation of efficient dense contact estimation on MOW~\cite{cao2021reconstructing} dataset.} MLLM inference efficiency is computed per sample.}
\begin{tabular}{lccc}
\toprule
 & \multicolumn{1}{c}{Contact accuracy} 
 & \multicolumn{2}{c}{MLLM inference efficiency} \\
Methods & F1-Score~$\uparrow$ 
        & \# of output tokens~$\downarrow$ & Cost~$\downarrow$ \\ 
\midrule 
w/o part cond. & 0.522 & 4,525 & \$0.136 \\ 
w/ part cond.~(Ours) & \textbf{0.526} & \textbf{3,588} & \textbf{\$0.108} \\
\bottomrule
\end{tabular}
\label{tab:abl_efficient_dense}
\end{table}

\noindent\textbf{Effectiveness of multi-stage contact reasoning.}
As shown in Table~\ref{tab:abl_multi_stage}, using only dense vertex-level stage results in low precision (0.382) and extremely high recall (0.944), indicating overly broad estimation.
Introducing part-level stage substantially improves precision from 0.382 to 0.460. 
Incorporating free-form reasoning stage results in the best overall F1-score of 0.526, demonstrating its role in providing global interaction context.

\noindent\textbf{Effectiveness of efficient dense contact estimation.}
As shown in Table~\ref{tab:abl_efficient_dense}, part conditioning improves F1-score by 0.8\%, while reducing the number of output tokens by 20.7\% and the cost per sample.
By restricting dense contact estimation to the selected contact parts, part conditioning reduces inference cost while maintaining contact accuracy.

\begin{table*}[pos=t]
\centering
\setlength{\tabcolsep}{4pt}
\caption{\textbf{Comparison of various MLLMs for dense hand contact estimation on MOW~\cite{cao2021reconstructing} dataset.} MLLM inference efficiency is computed per sample.}
\begin{tabular}{llccccc} 
\toprule
 & & \multicolumn{3}{c}{Contact accuracy} & \multicolumn{2}{c}{MLLM inference efficiency} \\
Source & Methods & Precision~$\uparrow$ & Recall~$\uparrow$ & F1-Score~$\uparrow$ 
        & \# of output tokens~$\downarrow$ & Cost~$\downarrow$ \\ 
\midrule 

\multirow{7}{*}{Open}
& GLM 4.1V 9B~\cite{hong2025glm} & \textbf{0.543} & 0.200 & 0.264 & \underline{1,490} & -- \\

& Gemma 3 12B~\cite{team2025gemma} & 0.430 & 0.293 & 0.283 & 1,980 & -- \\

& InternVL3 8B~\cite{zhu2025internvl3}
& \underline{0.497} & 0.180 & 0.242 & \textbf{908} & -- \\

& InternVL3.5 8B~\cite{wang2025internvl3}
& 0.385 & 0.406 & 0.336 & 1,866 & -- \\

& Qwen3.5 9B~\cite{qwen3.5}
& 0.383 & 0.464 & 0.374 & 2,663 & -- \\

& LLaVA OneVision 2 8B~\cite{an2026llava} & 0.339 & \textbf{0.593} & \underline{0.378} & 4,461 & -- \\

& Ministral 3 14B~\cite{liu2026ministral} & 0.468 & \underline{0.550} & \textbf{0.461} & 2,535 & -- \\

\midrule

\multirow{4}{*}{Closed}
& Claude Sonnet 4.6~\cite{anthropic2026claude_sonnet46} 
& 0.444 & \underline{0.661} & \underline{0.488} & \underline{1,978} & \underline{\$0.067} \\

& Claude Opus 4.7~\cite{anthropic2026claude_opus_47} 
& \textbf{0.506} & 0.534 & 0.473 & 2,308 & \$0.148 \\

& GPT-5.4~\cite{openai2026gpt54} 
& \underline{0.504} & 0.564 & 0.487 & \textbf{1,567} & \textbf{\$0.024} \\ 

& GPT-5.5~\cite{openai2026gpt55} 
& 0.473 & \textbf{0.710} & \textbf{0.526} & 3,588 & \$0.108 \\

\bottomrule
\vspace{-0.5cm}
\end{tabular}
\label{tab:abl_mllm}
\end{table*}

\begin{table}[pos=t]
\centering
\setlength{\tabcolsep}{2pt}
\caption{\textbf{Comparison with SOTA methods of dense hand contact estimation on MOW~\cite{cao2021reconstructing} dataset.}}
\begin{tabular}{lccc} \toprule
Methods & Precision~$\uparrow$ & Recall~$\uparrow$ & F1-Score~$\uparrow$ \\ 
\midrule 
POSA~\cite{hassan2021populating} & 0.134 & 0.128 & 0.101 \\ 
BSTRO~\cite{huang2022capturing} & 0.204 & 0.126 & 0.112 \\ 
DECO~\cite{tripathi2023deco} & 0.246 & 0.235 & 0.197 \\ 
HACO~\cite{jung2025learning} & \textbf{0.525} & \underline{0.607} & \underline{0.522} \\ 
\midrule
ContactPrompt~(Ours) & \underline{0.473} & \textbf{0.710} & \textbf{0.526} \\
\bottomrule
\end{tabular}
\label{tab:sota_hand_cont}
\end{table}

\begin{table}[pos=t]
\centering
\scriptsize
\setlength{\tabcolsep}{2pt}
\caption{\textbf{Comparison with SOTA method of dense hand contact estimation on various datasets of HIC~\cite{tzionas2016capturing}, RICH~\cite{huang2022capturing}, Hi4D~\cite{yin2023hi4d}.} For RICH and Hi4D, we evaluate on uniformly sampled subsets of their test sets.}
\begin{tabular}{llccc}
\toprule
Test dataset & Methods & Precision $\uparrow$ & Recall $\uparrow$ & F1-Score $\uparrow$ \\
\midrule
\multirow{2}{*}{HIC~\cite{tzionas2016capturing}} 
& HACO~\cite{jung2025learning} & 0.216 & 0.409 & 0.263 \\
& ContactPrompt~(Ours) & \textbf{0.284} & \textbf{0.516} & \textbf{0.350} \\
\midrule
\multirow{2}{*}{RICH~\cite{huang2022capturing}} 
& HACO~\cite{jung2025learning} & \textbf{0.774} & \textbf{0.923} & \textbf{0.814} \\
& ContactPrompt~(Ours) & 0.621 & 0.855 & 0.702 \\
\midrule
\multirow{2}{*}{Hi4D~\cite{yin2023hi4d}} 
& HACO~\cite{jung2025learning} & \textbf{0.677} & \textbf{0.698} & \textbf{0.630} \\
& ContactPrompt~(Ours) & 0.610 & 0.601 & 0.562 \\
\bottomrule
\end{tabular}
\label{tab:sota_all_dataset}
\end{table}

\begin{figure}[pos=t]
\begin{center}
\includegraphics[width=1.0\linewidth]{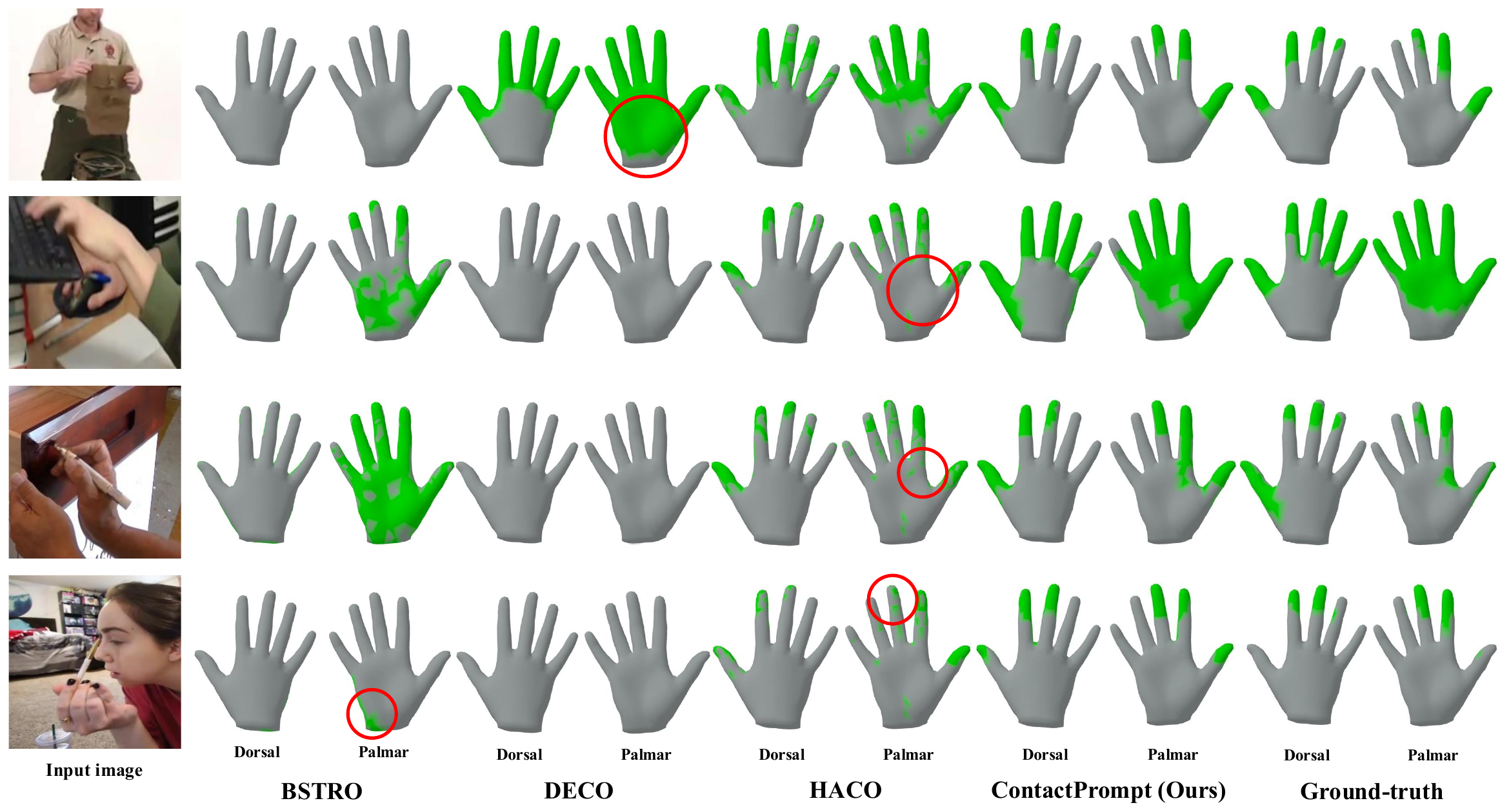}
\end{center}
\vspace{-3.5mm}
\caption{\textbf{Qualitative comparison of dense hand contact estimation with BSTRO~\cite{huang2022capturing}, DECO~\cite{tripathi2023deco}, HACO~\cite{jung2025learning} on MOW~\cite{cao2021reconstructing} dataset.} We highlight representative regions where ContactPrompt provides more accurate contact predictions with red circles.}
\label{fig:sota_qual_hand_contact}
\vspace{-0.5cm}
\end{figure}

\subsection{Comparison with state-of-the-art methods}

\noindent\textbf{Quantitative results.}
Table~\ref{tab:sota_hand_cont} presents a comparison between our method and previous methods, including POSA \cite{hassan2021populating}, BSTRO \cite{huang2022capturing}, DECO \cite{tripathi2023deco}, and HACO \cite{jung2025learning}, on the MOW \cite{cao2021reconstructing} dataset.
Our method achieves the best F1-score of 0.526, achieving performance comparable to the previous state-of-the-art method (HACO) in overall contact estimation, even without any task-specific training.
This result demonstrates that our training-free approach with MLLMs is effective for dense hand contact estimation.
We further compare ContactPrompt with HACO on additional hand contact datasets in Table~\ref{tab:sota_all_dataset}. 
Without any task-specific training, ContactPrompt outperforms HACO on the HIC dataset. 
On the RICH and Hi4D datasets, ContactPrompt remains competitive when compared with HACO.
Overall, ContactPrompt demonstrates strong performance across diverse hand interaction datasets without task-specific training.

\noindent\textbf{Qualitative results.}
Figure~\ref{fig:sota_qual_hand_contact} presents qualitative comparisons with BSTRO~\cite{huang2022capturing}, DECO~\cite{tripathi2023deco}, and HACO~\cite{jung2025learning} on the MOW~\cite{cao2021reconstructing} dataset. 
ContactPrompt produces more accurate contact predictions than previous methods. 
For example, the first row shows accurate contact at the thumb, index, and middle fingers, while HACO overestimates finger-base contact. 
The second row demonstrates that ContactPrompt accurately recovers both finger and palmar contact, whereas HACO misses the palmar region. 
In the third row, ContactPrompt correctly identifies the support region between the index and middle finger, while HACO only sparsely predicts this region as contact. 
Overall, these examples illustrate that ContactPrompt can produce spatially precise contact predictions across diverse interaction scenarios.

\noindent\textbf{Comparison of various MLLMs.}
In Table~\ref{tab:abl_mllm}, we compare the performance of different MLLMs for dense hand contact estimation.
Among the evaluated closed-source MLLMs, GPT-5.5 achieves the highest F1-score of 0.526. 
Among open-source MLLMs, Ministral 3 14B achieves the highest F1-score of 0.461.
Overall, ContactPrompt can be applied across a diverse range of open- and closed-source MLLMs, not restricted to a single proprietary MLLM.
\section{Limitations and societal impacts}
\label{sec:limitations}

Despite its effectiveness, ContactPrompt has several limitations.
First, the method relies on MLLMs, which are typically more computationally expensive than task-specific models, requiring an average inference time of 104.4 seconds per sample with GPT-5.5 on the MOW dataset.
Second, closed-source MLLMs depend on external MLLM APIs, which may incur costs and limit reproducibility due to potential changes in model behavior.
In terms of societal impacts, ContactPrompt can benefit applications involving hand interactions, such as AR/VR and robotics. However, it may also be misused in human monitoring scenarios, and we therefore discourage its misuse in such cases.
\section{Conclusion}
We propose ContactPrompt, a training-free, zero-shot approach for dense hand contact estimation using multi-modal large language models (MLLMs). 
To effectively encode 3D hand geometry for MLLMs, we introduce a detailed hand-part segmentation and a part-wise vertex-grid representation. 
For accurate and efficient dense contact prediction, we propose a multi-stage contact reasoning framework with part conditioning. 
Our method achieves performance comparable to the previous state-of-the-art method without requiring training on dense hand contact datasets.

\section*{Acknowledgements}
This work was supported in part by the IITP grants [No. RS-2021-II211343, Artificial Intelligence Graduate School Program (Seoul National University), No. RS-2025-02303870, No. 2022-0-00156] funded by the Korea government (MSIT).

\section*{Data availability}
The datasets used in this study are publicly available. The code for ContactPrompt is publicly available at \url{https://github.com/dqj5182/ContactPrompt_RELEASE}.

\section*{Declaration of competing interest}

The authors declare that they have no known competing financial interests or personal relationships that could have appeared to influence the work reported in this paper.

\printcredits

\bibliographystyle{cas-model2-names}
\bibliography{main}

\end{document}